\documentclass[
    textheight=9in,
    textwidth=5.6in,
    top=1in,
    headheight=12pt,
    headsep=25pt,
    footskip=30pt
  ]{article}

\usepackage[left=1in, right=1in, bottom=1in, top=1in]{geometry}

\usepackage[round]{natbib}
\PassOptionsToPackage{numbers, compress}{natbib}

\usepackage[utf8]{inputenc} %
\usepackage[T1]{fontenc}    %
\usepackage{hyperref}       %
\usepackage{url}            %
\usepackage{booktabs}       %
\usepackage{amsfonts}       %
\usepackage{nicefrac}       %
\usepackage{microtype}      %
\usepackage{xcolor}         %
\usepackage{graphicx}
\usepackage{algorithm}
\usepackage{algorithmic}
\usepackage{longtable}
\usepackage{amssymb}
\usepackage{mathtools}
\usepackage{amsthm}
\usepackage{multirow}
\usepackage{mdframed}
\usepackage{array} 
\usepackage{subcaption}
\usepackage{caption}

\usepackage{authblk}

\usepackage{listings}
\definecolor{codegreen}{rgb}{0,0.6,0}
\definecolor{codegray}{rgb}{0.5,0.5,0.5}
\definecolor{codepurple}{rgb}{0.58,0,0.82}
\definecolor{backcolour}{rgb}{0.95,0.95,0.92}

\lstdefinestyle{mystyle}{
    backgroundcolor=\color{backcolour},   
    commentstyle=\color{codegreen},
    keywordstyle=\color{magenta},
    numberstyle=\tiny\color{codegray},
    stringstyle=\color{codepurple},
    basicstyle=\ttfamily\footnotesize,
    breakatwhitespace=false,         
    breaklines=true,                 
    captionpos=b,                    
    keepspaces=true,                 
    numbers=none,                    
    numbersep=5pt,                  
    showspaces=false,                
    showstringspaces=false,
    showtabs=false,                  
    tabsize=2
}

\usepackage{amsmath,amsfonts,bm,amsthm,amssymb,mathrsfs,algorithm,algorithmic,url,xcolor,mathtools}
\usepackage{booktabs,enumitem,multicol,caption,subcaption,pifont,tikz,listings,wrapfig,lipsum,circledsteps,arydshln,hyperref,multirow}
\usepackage{relsize,footnote,makecell,graphicx,microtype,thmtools} % ulem
\usepackage[normalem]{ulem}

\usepackage[capitalize]{cleveref}

\hypersetup{
    colorlinks,
    linkcolor={blue!50!black},
    citecolor={blue!50!black},
    urlcolor={red!50!black}
}

\theoremstyle{plain}

\theoremstyle{definition}

\theoremstyle{remark}

\newtheorem*{theorem*}{Theorem}

\def\Secref#1{Section~\ref{#1}}
\def\eqref#1{(\ref{#1})}
\def\eqref#1{Eq.~(\ref{#1})}
\def\1{\bm{1}}

\DeclareMathAlphabet{\mathsfit}{\encodingdefault}{\sfdefault}{m}{sl}
\SetMathAlphabet{\mathsfit}{bold}{\encodingdefault}{\sfdefault}{bx}{n}

\def\gB{{\mathcal{B}}}

\def\gD{{\mathcal{D}}}

\def\gH{{\mathcal{H}}}

\def\gL{{\mathcal{L}}}
\def\gM{{\mathcal{M}}}

\def\gR{{\mathcal{R}}}

\def\sA{{\mathbb{A}}}

\def\sS{{\mathbb{S}}}

\newcommand{\E}{\mathbb{E}}

\newcommand{\R}{\mathbb{R}}

\newcommand{\br}[1]{\left(#1\right)}
\newcommand{\sbr}[1]{\left[#1\right]}
\newcommand{\given}{\,|\,}

\newcommand{\ie}{\textit{i.e.}}
\newcommand{\eg}{\textit{e.g.}}

\newcommand{\myparagraph}[1]{\textbf{#1~~}}

\definecolor{brickred}{rgb}{0.8, 0.25, 0.33}
\definecolor{darkspringgreen}{rgb}{0.09, 0.45, 0.27}
\definecolor{applegreen}{rgb}{0.55, 0.71, 0.0}
\definecolor{brightmaroon}{rgb}{0.76, 0.13, 0.28}
\definecolor{burgundy}{rgb}{0.5, 0.0, 0.13}
\newcommand{\pisft}{\pi_{\mathrm{SFT}}}

\newcommand{\cent}{c_{\mathrm{ent}}}

\title{\Large \textbf{Segmenting Text and Learning Their Rewards for Improved RLHF in Language Model}}

\author[1,2,\footnote{Equal contribution. $^\dagger$Corresponding Author. This work was done during an internship at Microsoft.}]{Yueqin Yin}
\author[1,*]{Shentao Yang}
\author[2,$\dagger$]{Yujia Xie}
\author[2,$\dagger$]{Ziyi Yang}
\author[2]{Yuting Sun}
\author[2]{\\Hany Awadalla}
\author[2]{Weizhu Chen}
\author[1,$\dagger$]{Mingyuan Zhou}
\affil[ ]{\texttt{yueqin.yin@utexas.edu, shentao.yang@mccombs.utexas.edu}}
\affil[ ]{\texttt{\{yujiaxie, ziyiyang, yutingsun, hanyh, wzchen\}@microsoft.com}}
\affil[ ]{\texttt{mingyuan.zhou@mccombs.utexas.edu }\vspace{3mm}}

\affil[1]{The University of Texas at Austin}
\affil[2]{Microsoft}
\date{ }

\begin{document}

\maketitle

\begin{abstract}
Reinforcement learning from human feedback (RLHF) has been widely adopted to align language models (LMs) with human preference.
Prior RLHF works typically take a bandit formulation, which, though intuitive, ignores the sequential nature of LM generation and can suffer from the sparse reward issue.
While recent works propose dense token-level RLHF, treating each token as an action may be oversubtle to proper reward assignment. 
In this paper, we seek to get the best of both by training and utilizing a segment-level reward model, which assigns a reward to each semantically complete text segment that spans over a short sequence of tokens.
For reward learning, our method allows dynamic text segmentation and compatibility with standard sequence-preference datasets.
For effective RL-based LM training against segment reward, we generalize the classical scalar bandit reward normalizers into location-aware normalizer functions and interpolate the segment reward for further densification.
With these designs, our method performs competitively on three popular RLHF benchmarks for LM policy: AlpacaEval 2.0, Arena-Hard, and MT-Bench.
Ablation studies are conducted to further demonstrate our method.
\end{abstract}

\section{Introduction}\label{sec:intro}

To align language models \citep[LMs, \eg,][]{gpt42023,reid2024gemini} with human values, reinforcement learning \citep[RL,][]{rlintro2018} methods have been widely adopted to optimize the non-differentiable human preference, leading to the paradigm of reinforcement learning from human feedback \citep[RLHF,][]{instructgpt2022,bai2022constitutional}.
A prevailing approach in RLHF is to optimize the LMs by proximal policy optimization \citep[PPO,][]{schulman2017proximal} against a \textit{bandit} reward model learned from human preference data, with KL regularization towards a pre-specified target distribution to avoid over-optimization on the reward model \citep{ziegler2019fine,stiennon2020learning,castricato2022robust}.
While this bandit approach is easier for reward modeling and has achieved remarkable success, language generation is intrinsically sequential, rather than simultaneous.
Thus, from the view of optimizing human preference, assigning a bandit reward to entire text sequence induces the sparse reward (delayed feedback) issue \citep{andrychowicz2017hindsight,marbach2003approximate}, that often hurts RL-based LM training by increasing gradient variance and lowering sample efficiency \citep{takanobu2019guided,wang2020learning,sqltext2021,snell2022offline}.

As efforts to mitigate this sparse reward issue, prior works have developed methods to ``ground'' the sequence-level preference label into a dense token-level reward model \citep{yang2023preferencegrounded,zhong2024dpo}.
While a dense {per-token} reward signal reduces the optimization complexity \citep{laidlaw2023bridging}, each action, however, is then defined as a single token, \ie, a \textit{sub-word} that is finer-grained than a word, especially with the BPE-style tokenizers \citep{gage1994new,sennrich2015neural}.
For instance, Llama 3.1's tokenizer \citep{llama31} has tokens as \texttt{\{Brit, ce, cod, neo, redd,...\}} that have less clear semantic meaning \textit{per se} in any given context.
The contribution of those tokens to the text sequence will inevitably depend on later tokens, making reward/credit assignment harder, especially under the prevailing RLHF paradigm of implementing the reward model as an off-the-shelf decoder-only transformer \citep[\eg,][]{instructgpt2022,bai2022constitutional,menick2022teaching}.
Further, token-level reward implicitly assumes that the basic unit of a text sequence is \textit{token}, which may not follow linguistics, where a more meaningful decomposition of text may be \textit{phrase} (including \textit{word}) that can be more semantically complete and generally consists of a short sequence of tokens. 

To retain the optimization benefit of dense reward for RLHF, while mitigating its potential reward assignment issue and linguistic counter-intuition, in this paper, we seek to train and utilize a \textit{segment-level} reward model, which assigns a reward to each semantically meaningful segment of text sequence that constitutes a small number of (or just one) tokens.
With this construction, we define the action space in RLHF as ``text segment,'' interpolating between the finest ``per token'' and the coarsest ``full sequence'' and potentially getting the benefit of both worlds: easier RL-based LM training owing to denser feedback and more accurate training guidance by the semantic completeness of each action. 

Technically, we are motivated by prior works \citep{malinin2018predictive,li2024cascade} to implement a dynamic text sequence segmentation by thresholding the entropy of LM's predictive distributions, under the assumption that tokens within a semantically complete text segment can be more certainly predicted by prior tokens, while the start of a new segment is not \citep{wang2024my}. 
To allow training the segment-level reward model by the standard sequence-preference labels via Bradley-Terry \citep[BT,][]{bradley1952rank} loss, we differentiably aggregate segment rewards in a text sequence into a parametrized sequence evaluation.
The learned segment-level reward model is then utilized in PPO-based policy learning, where we observe the unsuitability of classical reward normalizers, \ie, the mean and standard deviation (std) of full sequence rewards.
We address this issue by generalizing the classical bandit normalizers of scalar mean and std into a mean and a std function that output the reward normalizers at arbitrary locations of the text sequence. 
In addition, we enhance PPO training by within-segment reward interpolation, which further densifies training signal and improves results.

We test our method on the performance of the PPO-trained LM policy. 
On three popular RLHF benchmarks for LM policy: AlpacaEval 2.0,
Arena-Hard, and MT-Bench, our method achieves competitive performance gain against both the classical bandit reward approach and recent token-level reward approach.
We conduct a wide array of ablation studies to verify our design choices and provide further insight into our method.

\section{Main Method}\label{sec:method}

\subsection{Notations and Background} \label{sec:notation}
In this section, we  will define generic notations, provide background on the classical bandit RLHF, and then discuss RL formulation of LM generation underlying recent efforts on dense-reward RLHF.

\myparagraph{Generic Notations.} 
Both reward modeling and policy learning require text prompt $x$ and the corresponding response $y$.
Reward model training turns the supervised fine-tuned model $\pisft(\cdot \given \cdot)$ (without the final unembedding layer) into a parametrized scalar-output model $r_\phi(\cdot,\cdot)$ with parameter $\phi$ that scores its input.
The LM policy $\pi_\theta$, parametrized by $\theta$, is then optimized against $r_\phi$\,.

\myparagraph{Bandit Reward Model Training.}
Reward model training assumes a dataset $\gD_{\mathrm{pref}} = \{(x, y^w, y^l)\}$ of prompt $x$ and the corresponding winning/chosen response $y^w$ and losing/rejected response $y^l$, where the label of winning/losing comes from human evaluation on the entire text sequence $y^w$ and $y^l$.
In the classical bandit RLHF, reward model $r_\phi$ is trained by the binary classification BT loss
\begin{equation}\label{eq:bandit_bt}\textstyle
    \gL_{\mathrm{bandit}}(\phi) = -\E_{(x, y^w, y^l) \sim \gD_{\mathrm{pref}}}\sbr{\log\sigma\br{r_\phi(x, y^w) - r_\phi(x, y^l)}} \,,
\end{equation}
where $\sigma(u) = 1/(1+\exp\br{-u})$  denotes the sigmoid function.

\myparagraph{PPO-based Bandit Policy Learning.}
In policy learning, a set $\gD_{\mathrm{pol}} = \{x\}$ of text prompts $x$ is given. 
The LM policy $\pi_\theta$ is trained to generate outputs on $\gD_{\mathrm{pol}}$ optimizing the bandit reward from $r_\phi$\,, with a  KL penalty towards $\pisft$ to avoid reward over-optimization.
Collectively, the objective is
\begin{equation}\label{eq:bandit_ppo}\textstyle
\max_\theta \E_{x \sim \gD_{\mathrm{pol}}, y\sim \pi_\theta(\cdot \given x)}\sbr{r_\phi(x,y) - \beta \times \log\br{\pi_\theta(y\given x) / \pisft(y \given x)}} \,,
\vspace{-.1em}
\end{equation}
where $\beta$ is the KL coefficient. In practice, for PPO's training stability, the value of $r_\phi(x, y)$ is de-mean and de-std normalized based on statistics calculated on a calibration dataset, \eg, $\gD_{\mathrm{pref}}$.

\myparagraph{RL Formulation of LM Generation.}
By its sequential nature, LM generation is formulated as a Markov Decision Process (MDP) $\gM = (\sS, \sA, P, \gR, \gamma)$ \citep{rlintro2018}.
Concretely, for state space $\sS$, the state at timestep $t$, $s_t$, consists of the prompt $x$ and all generated tokens so far $a_{<t} =: [a_0, \ldots, a_{t-1}]$ with $a_{<0}=:\varnothing$, \ie, $s_t =: [x, a_{<t}]$.
$\sA$ is the action space, where the action $a_t$ at step $t$ is a short-sequence/segment of tokens from the vocabulary in our segment-level reward/policy setting, whereas $a_t$ is a single token in the token-level reward/policy setting.
Transition function $P$ deterministically appends the newly sampled tokens after the previous ones, \ie, $s_{t+1} = [s_t, a_t] = [x, a_{\leq t}]$.
$r(s,a): \sS \times \sA \rightarrow \R$ scores the action choice (segment/token selection) $a$ at state/context $s$ and is typically substituted by the learned reward model $r_\phi$\,.
$\gamma \in [0,1]$ is the discount factor.

In what follows, we will focus on our segment-level reward/policy setting where each action $a_t \in \sA$ is a semantically complete text segment, consisting of a non-deterministic number of consecutive tokens.
The response $y$ for prompt $x$ then contains a variable number of segments/actions, generically denoted as 
$y=[a_0, \ldots, a_{T-1}]$ where $T$ is the number of segments in $y$ and varies across responses.
When necessary, we denote a single token in $y$ as $y_i$\, whose generation context is $[x, y_{<i}]$.

\begin{figure}[t]
\centering
\vspace{-1em}
\includegraphics[width=\textwidth]{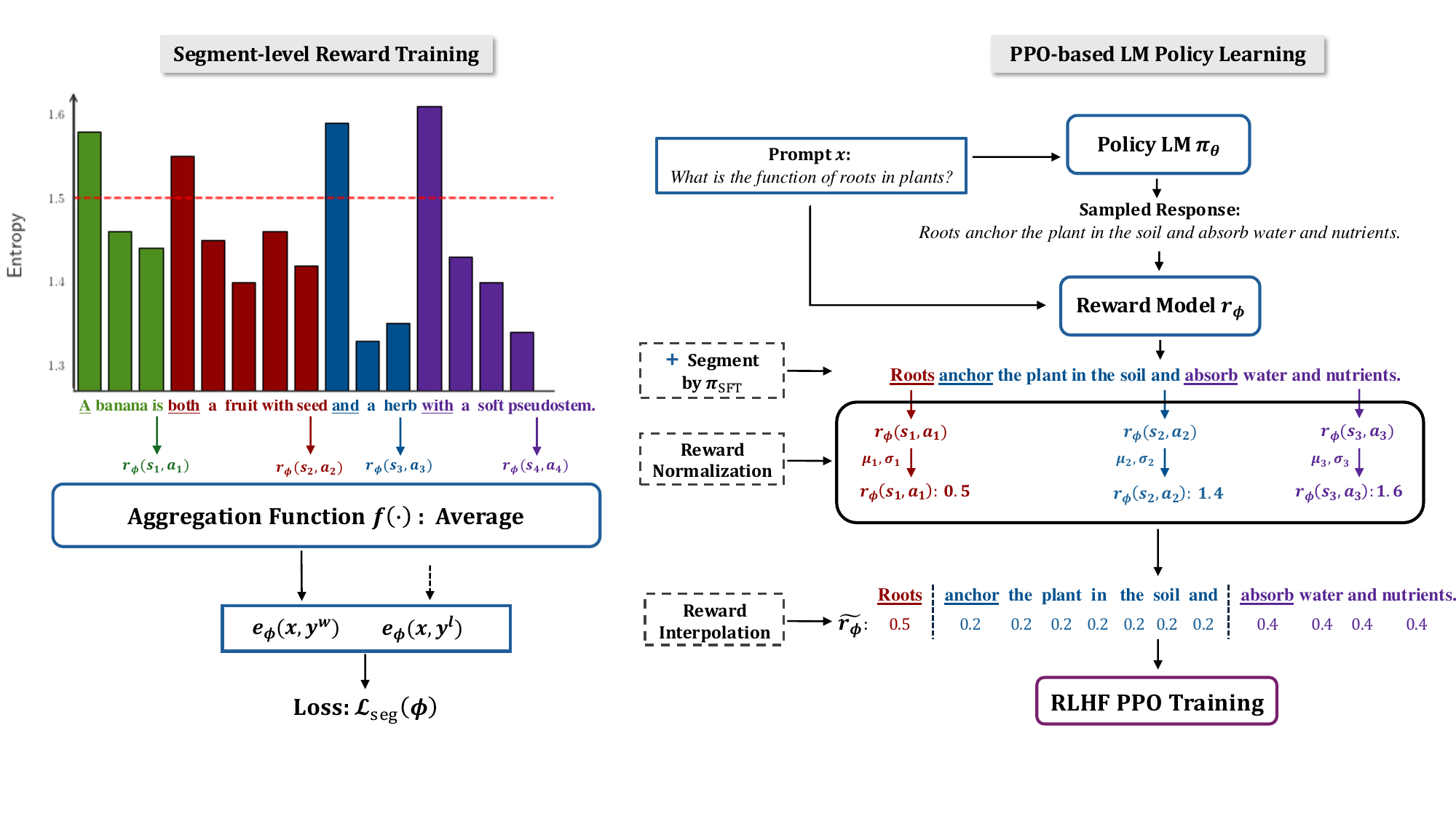}
\vspace{-3.6em}
\captionsetup{font=small}
\caption{
\small
Overview of training and utilizing our segment-level reward model. Numerics in the plot are artificial.
In the figure, each text segment has a different color, and its starting word is \uline{underscored}.
}
 \vspace{-1.25em}
\label{fig:method_overview}
\end{figure}

\cref{fig:method_overview} overviews key components in our method. A detailed algorithm box is deferred to \cref{sec:algo_box}.

\begin{figure}[t]
\centering
\includegraphics[width=\textwidth]{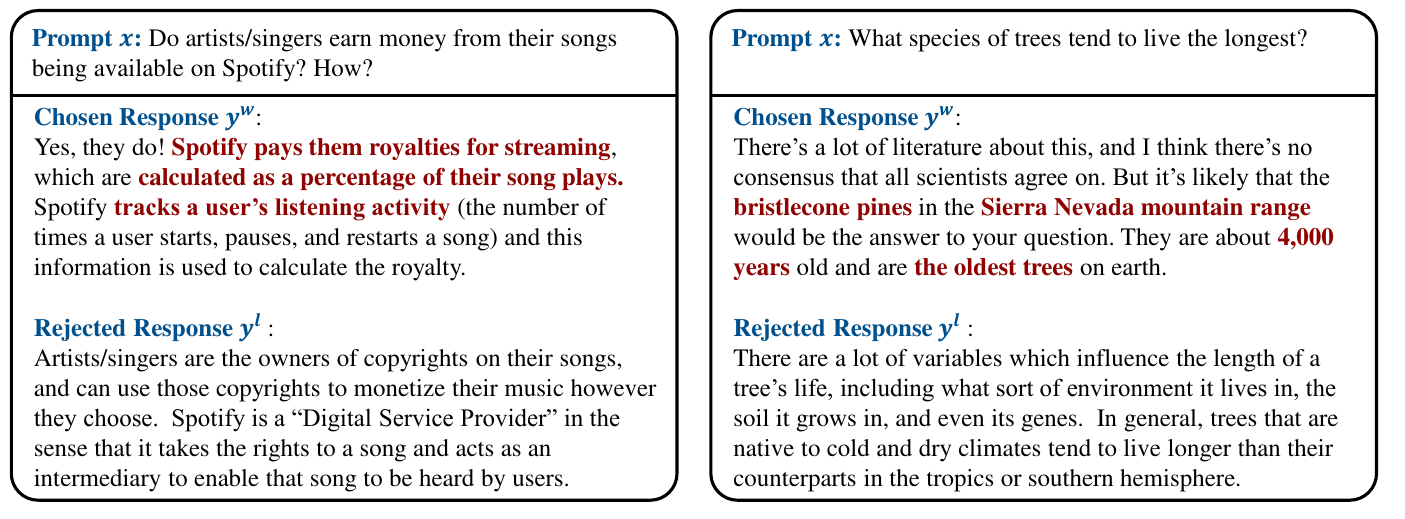}
\vspace{-1.55em}
\captionsetup{font=small}
\caption{
\small
Examples of data in our reward training dataset, motivating us to use Average as the aggregation function $f(\cdot)$.
It is clear that the average quality of the chosen responses is better than the rejected ones, as they contain some key phrases (highlighted in red) for responding to the given prompt.
We note that the choice of $f(\cdot)$ ought to be task and dataset specific, as discussed in \citet{yang2023preferencegrounded}.
}
 \vspace{-.95em}
\label{fig:reward_dataset_example}
\end{figure}

\subsection{Reward Model Training} \label{sec:method:reward_model}

\myparagraph{Overview.}
In training our segment-level reward model, we  follow the data assumption set forth in \cref{sec:notation}, 
where the dataset $\gD_{\mathrm{pref}} = \{(x, y^w, y^l)\}$ contains only binary sequence-level preference labels, without any process supervision \citep{uesato2022solving}.
The reward model $r_\phi(s_t, a_t)$ is configured to output a scalar reward for each text segment choice $a_t$ at the generation context $s_t$.
$r_\phi$ is trained such that its induced parameterized text sequence evaluations, aggregated from all segment-level rewards in the respective sequence, align with the preference labels in $\gD_{\mathrm{pref}}$.
This is inspired by the imitation learning literature \citep[\eg,][]{christiano2017deep,trex2019,drex2020} and prior token-level reward modeling in RLHF \citep{yang2023preferencegrounded}. 
Collectively, the BT loss for training our segment-level reward function $r_\phi$ is
\begin{equation}\label{eq:seg_bt}\textstyle
\resizebox{0.94\textwidth}{!}{%
    $
    \gL_{\mathrm{seg}}(\phi) = -\E_{(x, y^w, y^l) \sim \gD_{\mathrm{pref}}}[\log\sigma\br{e_\phi(x, y^w) - e_\phi(x, y^l)}], \forall\, y \in\{y^w, y^l\}, e_\phi(x, y) = f(\{r_\phi(s_t, a_t)\}_{a_t \in y}) \,,
$%
}
\end{equation}
where $e_\phi$ denotes the parameterized sequence evaluation induced by $r_\phi$, constructed by aggregating all segment-level rewards $\{r_\phi(s_t, a_t)\}_{a_t \in y}$ in the text sequence $y$ by a selected aggregation function $f(\cdot)$. Below, we discuss in detail the segmentation method for text sequence and the choice of $f(\cdot)$.

\myparagraph{Entropy-based Segmentation.} 
As discussed in \cref{sec:intro}, we intend to split the given text sequence $y \in \{y^w, y^l\}$ into semantically complete segments, so that the reward assignment to each action (segment) can be easier, especially under the common implementation of the reward model as an auto-regressive LM.
Recent works on LMs \citep[\eg,][]{li2024cascade,wang2024my} have noticed that tokens within a semantically complete segment can be more predictable by the corresponding generation context, since they are continuation of the designated semantics; while the starting token of a new segment is comparably less predictable, as its semantic binding with prior words is relatively weaker.
For auto-regressive LMs, the predictability of each token can be conveniently measured by the entropy of the next-token-prediction distribution from which the token is sampled \citep{malinin2018predictive}.
To make text sequence segmentation a one-time data pre-processing in reward model training, we choose to use the prediction distribution from the supervised fine-tuned model $\pisft$, from which the reward model is initialized before training.
With a selected entropy cutoff $\cent$, token $y_i$ starts a new segment if the Shannon entropy $\gH(\cdot)$ of $\pisft$'s predictive distribution of the $i$-th token surpasses $\cent$, \ie, $\gH(\pisft(\cdot \given x, y_{<i})) > \cent$, in which case $y_{i-1}$ ends the previous segment.

\myparagraph{Choice of the Aggregation Function $f(\cdot)$.} 
Aggregation function $f(\cdot)$ provides inductive bias on the relation between the quality of each segment/action and the preferability of overall text sequence.
Since $f$ probes into what kind of text sequences will be preferred, its selection should ideally be task and dataset specific, to avoid mis-specification and the subsequent unintended bias in reward model training.
While several designs have been proposed in literature \citep{christiano2017deep,kim2023preference,yang2023preferencegrounded}, after looking into the dataset, in our experiments, we select Average to differentiably highlight the better average quality of the chosen responses over the rejected ones.
With this choice of $f(\cdot)$, the parametrized sequence evaluation $e_\phi(x, y)$ in \cref{eq:seg_bt} is constructed as
\begin{equation}\label{eq:param_seq_eval}\textstyle
    e_\phi(x, y) = f(\{r_\phi(s_t, a_t)\}_{a_t \in y}) = \frac{1}{T}\sum_{t=0}^{T-1}r_\phi(s_t, a_t) \,.
\end{equation}
\cref{fig:reward_dataset_example} provides example data-points from the reward training dataset in our experiments (\cref{sec:exp}), supporting our choice of Average as the aggregation $f(\cdot)$.
Other datasets may require a different $f(\cdot)$.

\myparagraph{An Alternative Interpretation.}
Comparing our segment-level reward training loss \cref{eq:seg_bt} with the classical bandit loss \cref{eq:bandit_bt}, one may alternatively interpret $e_\phi$ and $f(\{r_\phi\})$ in \cref{eq:seg_bt} as a re-parametrization of the learned sequence-level feedback that differentiably aggregates the quality/contribution of each text segment, and thereby connects a denser evaluation $r_\phi$ of each semantically complete text segment with the information in ground-truth sequence-level preference label.

\subsection{PPO-based Policy Learning} \label{sec:method:ppo_policy}

\myparagraph{Overview.} 
\myparagraph{Overview.} 
In policy learning, we again follow the classical bandit setting in \cref{sec:notation} to optimize the LM policy $\pi_\theta$ on a given prompt set $\gD_{\mathrm{pol}} = \{x\}$.
But unlike the bandit objective in \cref{eq:bandit_ppo}, we adopt the full RL setting \citep{rlintro2018} to maximize $\pi_\theta$'s expected sum of per-segment/step rewards.
This enables directly plugging our segment-level reward model $r_\phi$ into most off-the-shelf RLHF PPO implementation.
With this, the policy learning objective for $\pi_\theta$ is
\begin{equation}\label{eq:seg_ppo_without_interpolate}\textstyle
\max_\theta \E_{x \sim \gD_{\mathrm{pol}}, y\sim \prod_{t=0}^{T-1} \pi_\theta(a_t \given s_t)}\sbr{\sum_{t=0}^{T-1} r_\phi(s_t,a_t) - \beta \times \log\br{\pi_\theta(y\given x) / \pisft(y \given x)}} \,,
\end{equation}
where again, each $a_t$ is a segment of tokens (chopped by $\pisft$), $s_t = [x, a_0, \ldots a_{t-1}]$ is the generation context at step $t$, and $y=[a_0, \ldots, a_{T-1}]$ is the response to prompt $x$ sampled from the learning $\pi_\theta$.

Recall from \cref{sec:notation} that the output values from the reward model $r_\phi$ need to be normalized for the stability of PPO training.
With our segment-level reward model, it is no longer suitable to normalize each $r_\phi(s_t, a_t)$ by the mean and std of entire sequences' rewards as in the bandit setting, since the latter may not be on a proper scale  and/or well-defined.
Further, the on-policy nature of PPO induces an extra complexity: each step of PPO samples new text sequences, whose total length, segment lengths, and segment locations are all stochastic and can differ from the reward  calibration dataset, \eg, $\gD_{\mathrm{pref}}$.
\cref{sec:more_on_reward_norm} provides an extended discussion on reward normalization in PPO-based LM training.
Below, we discuss our approach to construct the reward value normalizers, followed by interpolating the segment-level reward into per-token signal to helpfully provide an even denser training guidance.

\myparagraph{Location-aware Reward Normalizers via Regression.}
While the length of the sampled {response} $y$ and the lengths and locations of segments $\{a_t\}$ in $y$ are all stochastic, we know that each $a_t$ {is somewhere in $y$. Correspondingly,} % completes part of $y$.
each input $(s_t, a_t)$ to $r_\phi$ is linked to a {normalized location} $p \in (0,1]$ of $y$, and $p$ can be simply defined as $t/T$, where $t$ is the index of {the} segment $a_t$ in $y$, since PPO routine has fully sampled $y$.
On each {datapoint} in the calibration set, {normalized location} $p \in (0,1]$ again, with the linked segment-level reward available.
Across all {data points} in the calibration set, we construct a new dataset $\gD_{\mathrm{norm}} = \{(p, \mu_p, \sigma_p)\}$, where $p$ runs over all values of {normalized location} in the calibration set, $\mu_p$ and $\sigma_p$ respectively denote sample mean and sample std of all segment-level rewards corresponding to $p$ in the calibration set. 
With \(\gD_{\mathrm{norm}}\), we run a simple linear regression to estimate the relation between the log-transformed {normalized location} \(\log(p)\) and the mean/std of segment-level rewards at $p$. Specifically, the regression formula is given by:
\begin{equation}\label{eq:normalizer_reg}\textstyle
    \mathrm{Mean}(p) = w_\mu \times \log(p) + b_\mu, \; 
    \mathrm{Std}(p) = w_\sigma \times \log(p) + b_\sigma,
\end{equation}
where the independent variable is \(\log(p)\) and the regression coefficients \((w_\mu, b_\mu)\) and \((w_\sigma, b_\sigma)\) can be calculated in closed form.

Note that the classical bandit normalizers of the mean and std of full sequences' rewards correspond to evaluate $\mathrm{Mean}(p)$ and $\mathrm{Std}(p)$ at $p=1.0$.
In this regard, our mean and std functions in \cref{eq:normalizer_reg} generalize the classical scalar normalizers into location-aware functions able to output proper reward normalizers at an arbitrary {(normalized) location} $p$ of the text sequence.
With $\mathrm{Mean}(\cdot)$ and $\mathrm{Std}(\cdot)$ and the corresponding $p$, $r_\phi(s_t, a_t)$ is normalized by $r_\phi(s_t, a_t) \leftarrow (r_\phi(s_t, a_t) - \mathrm{Mean}(p))/\mathrm{Std}(p)$.

\myparagraph{Within-segment Reward Interpolation.} 
Depending on the specific tokenizer in use, we observed that semantically complete text segments may contain up to around twenty tokens.
The corresponding action space $\sA$ might still be  large and
the resulting segment-level design might not sufficiently address the sample inefficiency issue in the classical bandit RLHF and could again lead to inferior PPO-based RL training.
To further densify the RL training signal, we evenly split the segment-level reward $r_\phi(s_t, a_t)$ for a segment $a_t$ to each token $y_i \in a_t$.
This induces a token-level credit assignment that $\forall\, y_i \in a_t, \widetilde r_\phi([x, y_{<i}], y_i) = r_\phi(s_t, a_t) / |a_t|$, where $[x, y_{<i}]$ is  the generation context of token $y_i$.
$\widetilde r_\phi$ can then directly substitute $r_\phi$ in \cref{eq:seg_ppo_without_interpolate}, since $\sum_{t=0}^{T-1} r_\phi(s_t,a_t) = \sum_{t=0}^{T-1} (\sum_{ y_i \in a_t} r_\phi(s_t,a_t) / |a_t|)$. 

Note that \textit{$\widetilde r_\phi$ is still intrinsically segment level}, since all token selections $y_i$ within segment $a_t$ receive the same feedback, \ie, the average of segment-level reward $r_\phi(s_t, a_t)$.
This is in contrast to prior works on token-level reward models \citep{yang2023preferencegrounded,zhong2024dpo}, where each token selection is evaluated separately and thus their token-level feedback/rewards vary for each token.

\myparagraph{Summary.} 
With the learned segment-level reward model $r_\phi$ from \cref{sec:method:reward_model}, in PPO training of the LM policy $\pi_\theta$, we first normalize each $r_\phi(s_t, a_t)$ in the sampled sequence by the corresponding normalizers $\mathrm{Mean}(p)$ and $\mathrm{Std}(p)$.
Normalized segment-level rewards are then interpolated into the per-token feedback signal $\widetilde r_\phi$.
Finally, we plug $\widetilde r_\phi$ directly into an off-the-shelf RLHF PPO routine.

\section{Related Work} \label{sec:related_work}

\myparagraph{Reward Models in RLHF.}
In the classical RLHF paradigm, policy LM is optimized against a bandit reward model trained firstly by binary classification loss on the preference dataset, with KL penalty to a specified prior distribution to avoid reward over-optimization \citep{ziegler2019fine,stiennon2020learning,jaques2020human,bai2022training,instructgpt2022,castricato2022robust}. 
Under the same bandit formulation, recent works have enhanced the bandit reward model by directly modeling the probability of one response being preferred over the other \citep{jiang2023llm,zhao2023slic,liu2023statistical,dong2024rlhf} or factorizing human preference into multiple facets via multi-objective modeling \citep{touvron2023llama,wang2023helpsteer,wang2024helpsteer2,wang2024interpretable}.
Despite its popularity, from the angle of RL-based optimization of human preference captured by the reward model, such a bandit reward may lead to inferior training, due to the sparse reward issue intrinsic to the bandit formulation of LM generation and credit assignment \citep[\eg,][]{takanobu2019guided,sqltext2021}.

% papers on per-step dense reward RLHF
Viewing the weakness of bandit RLHF, efforts have been making to densify the reward signal for RLHF LM training.
\citet{yang2023preferencegrounded} and \citet{chan2024dense} train token-level reward models by the binary preference classification loss.
\citet{zhong2024dpo} and \citet{rafailov2024r} use an LM trained by DPO \citep{dpo2023} firstly for token-level reward assignment, which is later used in PPO training or search-based algorithms.
\citet{guo2023beyond}, \citet{cao2024drlc}, and \citet{yoon2024tlcr} assign continuous or fixed fine-grained rewards (\eg, $\pm 1$) by accessing 
an external powerful large LM or the oracle environmental reward; while \citet{chen2024improving} require the extra task and datasets of erroneous solution rewriting.
Apart from potential extra requirements, as discussed in \cref{sec:intro}, the semantic incompleteness of \textit{token} in text may challenge the efficacy of per-token credit assignment, especially with the prevailing implementation of reward model as a decoder-only transformer that cannot look ahead into later tokens.
In contrast, by defining \textit{text segment} as the basic unit of text sequence that can be semantically more complete than \textit{token},  our segment-level reward may provide more accurate guidance for RL-based LM training, while not losing the benefit of denser feedback.

Close to our segment-level reward, process reward models \citep[PRMs, \eg,][]{uesato2022solving,lightman2023let} in reasoning-alike tasks also assign a single reward value to a short sequence of tokens.
The training of PRMs, however, typically requires human annotation on each step of the reasoning-alike process. 
This may not be feasible in general text generation tasks, \eg, text summarization or dialog, where each step/text segment lacks clear human evaluation criterion while the full generations can be more easily compared or evaluated.
By contrast, as seen in \cref{sec:method}, our method is developed for the most basic yet general RLHF setting, where human preference is only manifested in a dataset of binary sequence-level preference.
And the dataset is collected from multiple sources and contains multiple forms of prompt-responses.

\myparagraph{Learning-from-preference.}
Learning-from-preference classically takes a two-stage approach where a reward model is first trained on a dataset of binary or multiple ranking via maximizing the choice model likelihood \citep{bradley1952rank,plackett1975analysis,luce2012individual}, before optimizing the RL/control policy against the learned reward model by RL algorithms \citep{akrour2011preference,akrour2012april,furnkranz2012preference}.
Earlier application in deep learning mainly focuses on relatively simple neural-network policy for robotics/control tasks \citep{christiano2017deep,ibarz2018reward,biyik2019green,trex2019,drex2020,Lee2021PEBBLEFI,shin2021offline,hejna2023few,hejna2023inverse}.
Implanting its success in robotics, in natural language generation, this two-stage learning-from-preference paradigm has been scaled up and popularized in the post-training stage to align LMs with specific human values, with applications ranging from text summarization \citep{ziegler2019fine,stiennon2020learning}, prompt generation \citep{yang2023preferencegrounded}, to (task-oriented) conversational agent \citep[\eg,][]{instructgpt2022,bai2022training,menick2022teaching,fantasticrewards2022,gpt42023}, and with the RL paradigm of both model free \citep{offlinetutorial2020,jointmatching2022} and model based \citep{sdmgan2022,wmbrl2022}.

To alleviate the complexity in fitting an explicit reward model, motivated by the theory of maximum-entropy control and RL \citep{ziebart2008maximum,ziebart2010modeling,Finn2016ACB}, direct preference optimization methods \citep[DPO, \eg,][]{dpo2023, tunstall2023zephyr, azar2023general, yuan2023rrhf, zhao2023slic,ethayarajh2023halos,yin2024relative} were recently proposed to directly train LMs on a preference dataset by using their log-density-ratio as the classification logit, which have been adapted to train text-to-image diffusion models \citep[\eg,][]{wallace2023diffusion,yangdense,li2024aligning,gu2024diffusion}.

In this paper, we contribute to the literature of learning-from-preference by re-thinking a suitable definition of action space in the RL formulation of LM generation and preference alignment.
Motivated by semantic completeness in linguistics, we define each action as ``text segment'', spanning across a small amount of tokens and interpolating between prior works' action space of either the finest ``per token'' or the coarsest ``full sequence''.
With this design, our method may benefit from both denser reward signal for easier RL-based LM training and the semantic completeness of each action for more accurate training guidance, as experimentally verified in \cref{sec:exp}.

\myparagraph{Training Signals for RL-based Language Model (LM) Training.}
In RL-based LM fine-tuning, a classical training signal for adapting LMs to the specific downstream task is the native trajectory-level downstream test metrics \citep[\eg,][]{ryang2012framework,ranzato2015sequence,rennie2017self,paulus2017deep,shu2021reward,quark2022}. 
This approach intrinsically uses a bandit formulation of LM generation that treats the entire generated sequence as a single action.
As discussed in \Secref{sec:intro}, ignoring the sequential nature of LM generation, this bandit training signal delays the feedback to each token/phrase selection, and can thus incur optimization difficulty \citep{sqltext2021,snell2022offline}.
With various forms of stronger data or compute requirements, task-specific per-step training signals have been proposed to mitigate this sparse reward issue.
Assuming abundant golden expert data for supervised (pre-)training,
\citet{shi2018toward} construct per-step reward via inverse RL \citep{russell1998learning};  
\citet{leakgan2018} use a hierarchical approach; 
\citet{yang2018unsupervised} learn LM discriminators; 
\citet{adversarialranking2017} and \citet{yu2017seqgan} use the expensive and high-variance Monte Carlo rollout to estimate per-step reward from a sequence-level adversarial reward function trained in the first place;
while \citet{le2022coderl} use some rule-based intermediate training signal derived from the oracle sequence-level evaluation, without explicitly learning per-step reward.

Similarly, in RLHF, to move forward from the classical bandit formulation, methods have recently been proposed to ground sparse preference labels into dense per-step feedback, with applications in task-oriented dialog systems \citep[\eg,][]{caspi2021,fantasticrewards2022} and variable-length text-sequence generation \citep{yang2023preferencegrounded}.
Our paper seeks to reconcile dense \textit{v.s.} sparse training signal in RLHF by distributing feedback to the level of semantically complete ``text segment'', interpolating between the densest ``token level'' and the sparsest ``sequence level'' and ideally getting the benefit of both worlds: easier RL training and accurate optimization signal. Meanwhile, as seen in \cref{sec:method}, our method adheres to the classical two-stage RLHF paradigm without requiring extra data or compute.

\myparagraph{Other LM Preference Alignment Methods.}
Apart from RL methods, strategies have been developed to align LMs with preference by adding external filters on top of the pretrained LMs, for safety checking the generations or the training texts \citep[\eg,][]{xu2020recipes}.
Vanilla maximum likelihood estimation has also been adopted for training LMs on curated datasets \citep{hancock2019learning,solaiman2021process,scheurer2022training}, or instruction fine-tuning LMs on massive highly-curated sets of tasks \citep{sanh2022multitask,chung2022scaling}.
With extra requirements on data, modelling, and/or compute, recent LM works also conduct preference alignment by formulating text generation as a constraint satisfaction problem on LM's generation distribution  \citep[\eg,][]{khalifa2021a,korbak2022reinforcement,go2023aligning}, or utilizing the preference dataset in LMs' pre-training stage \citep{korbak2023pretraining}.

In this paper, we seek to refine RL-based LM preference alignment by re-thinking the suitable action space in the RL formulation that allows both denser immediate feedback while not jeopardizing the feedback accuracy.
Our segment-level design is validated through numeric and example in \cref{sec:exp}.

\section{Experiments}\label{sec:exp}

\subsection{Experimental Setups and Implementation} \label{sec:exp_setup}

\myparagraph{Datasets.} 
For reward model training, we use the preference-700K dataset\footnote{\url{https://huggingface.co/datasets/hendrydong/preference_700K}}, 
which is a diverse collection of open-source preference datasets, such as HH-RLHF \citep{bai2022training}, 
Stanford Human Preferences Dataset (SHP) \citep{ethayarajh2022understanding}, 
and HelpSteer \citep{wang2023helpsteer}. 
PPO-based LM policy training is conducted on the Ultrafeedback dataset \citep{cui2023ultrafeedback}, from which we only use the prompts to sample responses during the PPO training routine.

\myparagraph{Benchmarks and Evaluations.}
The (PPO-trained) LM policy is evaluated on three popular open-ended instruction-following benchmarks: AlpacaEval 2.0 \citep{li2023alpacaeval}, Arena-Hard \citep{li2024live}, and MT-Bench \citep{zheng2023judging}, where GPT-4o is used as the judge. 
 Our reported scores follow each benchmark's default protocol.
 AlpacaEval 2.0 consists of 805 test cases, on which we report the length control win rate (\texttt{LC}), raw win rate (\texttt{WR}), and the average response length in number of characters (\texttt{\# char}).
 The \texttt{LC} metric is specifically designed to be robust against model verbosity.
We follow the convention \citep[\eg,][]{dong2024rlhf} to employ \texttt{alpaca\_eval\_gpt4\_turbo\_fn} as the annotator for AlpacaEval 2.0. 
 Arena-Hard consists of 500 challenging user queries, and we report both the win rate (\texttt{WR}) against the reference model GPT-4-0314 and the average number of tokens in the responses (\texttt{\# token}).
 MT-Bench is a multi-turn dialogue dataset that includes 80 test cases. We report the average MT-Bench scores for two dialogue turns, using GPT-4o as the judge model. 

\myparagraph{Implementation.}
We implement our method onto the open-sourced 3.8B Phi3-mini Instruct \citep{abdin2024phi} and the SFT checkpoint of Phi3.1-mini Instruct, as well as the popular SFT checkpoint of Llama-3-8B \citep{llama31} released by RLHFlow \citep{dong2024rlhf}\footnote{\url{https://huggingface.co/RLHFlow/LLaMA3-SFT-v2}}. The backbone model is used as the starting points of both reward model training and PPO-based LM policy learning, in the latter initializing the models for value function, learning policy, and reference policy.   
Our implementation is built upon the open-source RLHF framework \href{https://github.com/OpenRLHF/OpenRLHF}{OpenRLHF} \citep{hu2024openrlhf}.
We maximally follow the default hyperparameters in OpenRLHF.
For example, in PPO training, we retain the default KL coefficient value $\beta = 0.01$.
% In both reward model training and LM policy learning, we train the models for two epoch, \ie, two passes through the dataset, using entropy cutoff $\cent = 1.75$.
In the main experimental results, we train the reward model for one epoch and the PPO model for two epochs, \ie, two passes through the dataset, using entropy cutoff $\cent = 1.75$.
\cref{sec:exp_abla} presents ablation studies on the choice of $\cent$.
Due to space limit, we defer further implementation details to \cref{sec:exp_details}. 
For reproducibility, our source code and model checkpoints are  \href{https://github.com/yinyueqin/DenseRewardRLHF-PPO}{released}.

\subsection{Main Experimental Comparisons} \label{sec:exp_main_results}

\myparagraph{Baselines.}
To demonstrate our unique consideration of RLHF's action space, in the main experiment, we compare our design of segment-level action space with the coarsest bandit/sequence-level action space, the coarser sentence-level  space, and the finest token-level space, in terms of  performance of the PPO-trained LM policy.
For PPO training, a corresponding reward model is first trained under the specified action definition.
The sentence-level models are implemented by splitting the text sequences using sentence splitters \texttt{\{".", "!", "?", "\textbackslash n", ";", "…", ",", ":"\}} and/or their foreign language equivalents.

To further illustrate our segment-level reward model and denser segment-level reward assignment, we additionally compare with two hybrid approaches: \textbf{\textit{(A)}} using the bandit reward model for segment-level reward assignment in the PPO training (``Bandit as Segment''); and \textbf{\textit{(B)}} using the segment-level reward model only for bandit reward assignment in the PPO training (``Segment as Bandit''), where the bandit reward is implemented by the parametrized sequence evaluation $e_\phi$ in \cref{eq:param_seq_eval}.
For all baselines, we follow the standard training receipts and tune them to the extent of ensuring a fair comparison.

\begin{table}[t]
\captionsetup{font=small}
\caption{
\small
Performance comparison among different action definitions on PPO-trained LM policy, with the backbone model being Phi3-mini Instruct.
\# \{char, token\} measures the average response length in the benchmark tests.
Highest value of each column is in bold.
} 
\label{table:diff_seg_method}
\centering 
\vspace{-1em}
\begin{tabular}{@{}lccccccccccc@{}}
\toprule
\multirow{2}{*}{\begin{tabular}[c]{@{}l@{}}Action \\ Definition\end{tabular}} &  & \multicolumn{3}{c}{AlpacaEval 2.0}              &  & \multicolumn{2}{c}{Arena-Hard} &  & MT-Bench      \\  \cmidrule(lr){3-5}  \cmidrule(lr){7-8}  \cmidrule(lr){10-10}
\multicolumn{1}{c}{}                                                                              &  & LC(\%)         & WR(\%)         & \# char       &  & WR\%          & \# token       &  & GPT-4o        \\ \midrule
Phi3-mini Instruct                                                                                      &  & 18.89          & 14.41          & 1473          &  & 25.1          & 490            &  & 7.33           \\\midrule
Bandit (Sequence)                                                                                 &  & 27.05          & 29.07          & 2164          &  & 31.3          & 623            &  & 7.46          \\
Sentence                                                                                          &  & 25.56          & 32.92          & \textbf{2626} &  & 32.8          & \textbf{671}   &  & 7.51          \\
Token                                                                                             &  & 27.82          & 26.46          & 1940          &  & 27.2          & 533            &  & 7.58          \\
\midrule
Segment (\textbf{Ours})                                                        &  & \textbf{31.05} & \textbf{34.53} & 2257          &  & \textbf{34.0}   & 593            &  & \textbf{7.65} \\\midrule
Bandit as Segment                                                                                 &  & 14.39          & 6.46           & 691           &  & 11.1          & 308            &  & 6.61          \\
Segment as Bandit                                                                                 &  & 27.15          & 28.20          & 2079          &  & 30.9          & 620            &  & 7.38         \\ \bottomrule
\end{tabular}
\vspace{-1em}
\end{table}

\myparagraph{Results.} 
\cref{table:diff_seg_method} compares the performance of our PPO-trained LM policy with other definitions of RLHF's action space, as well as the two hybrid approaches, under the backbone model of Phi3-mini Instruct.
Our key findings are summarized as follows.

\textit{(1) Our segment-level approach improves RLHF training while not suffering from length hacking.~~}
As seen in \cref{table:diff_seg_method}, our LM policy performs better than the baselines across all three RLHF policy evaluation benchmarks: AlpacaEval 2.0, Arena-Hard, and MT-Bench.
Notably, our model's average response length on AlpacaEval 2.0 and Arena-Hard is not significantly larger than the baseline models', in contrast to the LM policy from the sentence-level action space.
Together, these results manifest the merit of our segment-level approach in truly improving the quality of the generated responses while not cheating the benchmark evaluations by response-length hacking \citep{dubois2024length}.
As discussed in \cref{sec:intro}, we attribute the gain of our segment-level approach over the baselines to simultaneously achieving both denser reward signals for PPO-based RLHF training and more accurate reward assignment by the design of semantically complete action space.

\textit{(2) Not all finer action spaces can help RLHF training over the classical bandit formulation.~~}
Apart from our denser segment-level approach, in \cref{table:diff_seg_method}, we see  that the other two finer action space specifications: per-sentence and per-token, both fail to generally improve over the classical design of bandit/sequence-level action space, especially on AlpacaEval 2.0 and Arena-Hard.
This provides an extra verification to our receipt of segment-level action space for RLHF PPO training, which provides a more break-down action definition compared to the coarser sentence level, and more complete semantics compared to the token-level design.

\textit{(3) A segment-level reward model is necessary for segment-level reward assignment, vice versa.~~}
One may wonder if we can use the classical bandit reward model to assign segment-level reward in the PPO training. 
As shown by the results of ``Bandit as Segment'' in \cref{table:diff_seg_method}, this approach performs significantly worse than the original pure bandit, which in turn under-performs our segment-level design.
These comparisons justify the necessity to train a segment-level reward model for segment-level reward assignment.
Conversely, using our segment-level reward model to provide only bandit feedback in PPO training (``Segment as Bandit'') leads to slight performance degradation over pure bandit design.
Compared with our main results, we see that ``Segment as Bandit'' does not fully benefit from our proposal of a (consistent) segment-level action space.
Its weaker results again highlight the gain of denser reward assignment in PPO-based RLHF training.

\begin{table}[t]
\captionsetup{font=small}
\caption{
\small
Performance comparison among different action definitions on PPO-trained LM policy, with the backbone model being the  3.8B SFT checkpoint of Phi3.1-mini Instruct. Table format follows \cref{table:diff_seg_method}.
} 
\label{table:action_definitions}
\centering 
\vspace{-1em}
\begin{tabular}{@{}lcccccccc@{}}
\toprule
\multirow{2}{*}{\begin{tabular}[c]{@{}l@{}}Action \\ Definition\end{tabular}} & 
\multicolumn{3}{c}{AlpacaEval 2.0} & 
& \multicolumn{2}{c}{Arena-Hard} & 
& MT-Bench \\  
\cmidrule(l){2-4} \cmidrule(l){6-7} \cmidrule(l){9-9}
 & LC (\%) & WR (\%) & \# char & & WR (\%) & \# token & & GPT-4o \\ 
\midrule
Phi3.1-mini SFT & 14.93 & 10.19 & 1271 & & 14.5 & 476 & & 7.00 \\ 
Bandit (Sequence) & 19.39 & 14.78 & 1542 & & 19.5 & 524 & & 7.26 \\ 
Token & 22.48 & 19.25 & 1687 & & 23.2 & 525 & & 7.43 \\ \midrule
Segment (\textbf{Ours}) & \textbf{26.19} & \textbf{23.85} & \textbf{1795} & & \textbf{28.5} & \textbf{585} & & \textbf{7.49} \\ 
\bottomrule
\end{tabular}
\end{table}

\begin{table}[t]
\captionsetup{font=small}
\caption{
\small
Performance comparison among different action definitions on PPO-trained LM policy, with the backbone model being the 8B SFT checkpoint of Llama-3 released by RLHFlow. Table format follows \cref{table:diff_seg_method}.
} 
\label{table:action_definitions_llama3}
\centering 
\vspace{-1em}
\begin{tabular}{@{}lcccccccc@{}}
\toprule
\multirow{2}{*}{\begin{tabular}[c]{@{}l@{}}Action \\ Definition\end{tabular}} & 
\multicolumn{3}{c}{AlpacaEval 2.0} & 
& \multicolumn{2}{c}{Arena-Hard} & 
& MT-Bench \\  
\cmidrule(l){2-4} \cmidrule(l){6-7} \cmidrule(l){9-9}
 & LC (\%) & WR (\%) & \# char & & WR (\%) & \# token & & GPT-4o \\ 
\midrule
Llama-3-8B-SFT & 16.31 & 9.50 & 1221 & & 10.4 & 469 & & 6.82 \\ 
Bandit (Sequence) & 21.20 & 20.99 & 2218 & & 18.7 & 513 & & 7.11 \\ 
Token & 23.84 & 20.87 & 1744 & & 26.0 & \textbf{622} & & 7.13 \\ \midrule
Segment (\textbf{Ours}) & \textbf{25.11} & \textbf{28.57} & \textbf{2264} & & \textbf{30.4} & {616} & & \textbf{7.15} \\ 
\bottomrule
\end{tabular}
% }
% \vspace{-1em}
\end{table}

\textit{(4) The benefit of segment-level design extends to SFT model and the larger 8B model.~~}
We swap the backbone model to the SFT checkpoint of Phi3.1-mini Instruct, as shown in \cref{table:action_definitions}, and to the larger 8B SFT checkpoint of Llama-3, as presented in \cref{table:action_definitions_llama3}.

It is clear the the gain of our segment-level action space over the prior bandit and token-level design is not scoped within the already DPO'ed Phi3-mini Instruct.
Rather, our advantage extends to both the SFT checkpoint of Phi3.1-mini Instruct and the larger Llama-3-8B-SFT, which verifies the value and versatility of our method in the practical post-training pipeline.

\begin{figure}[t]
\centering
% \vspace{-1em}
\includegraphics[width=\textwidth]{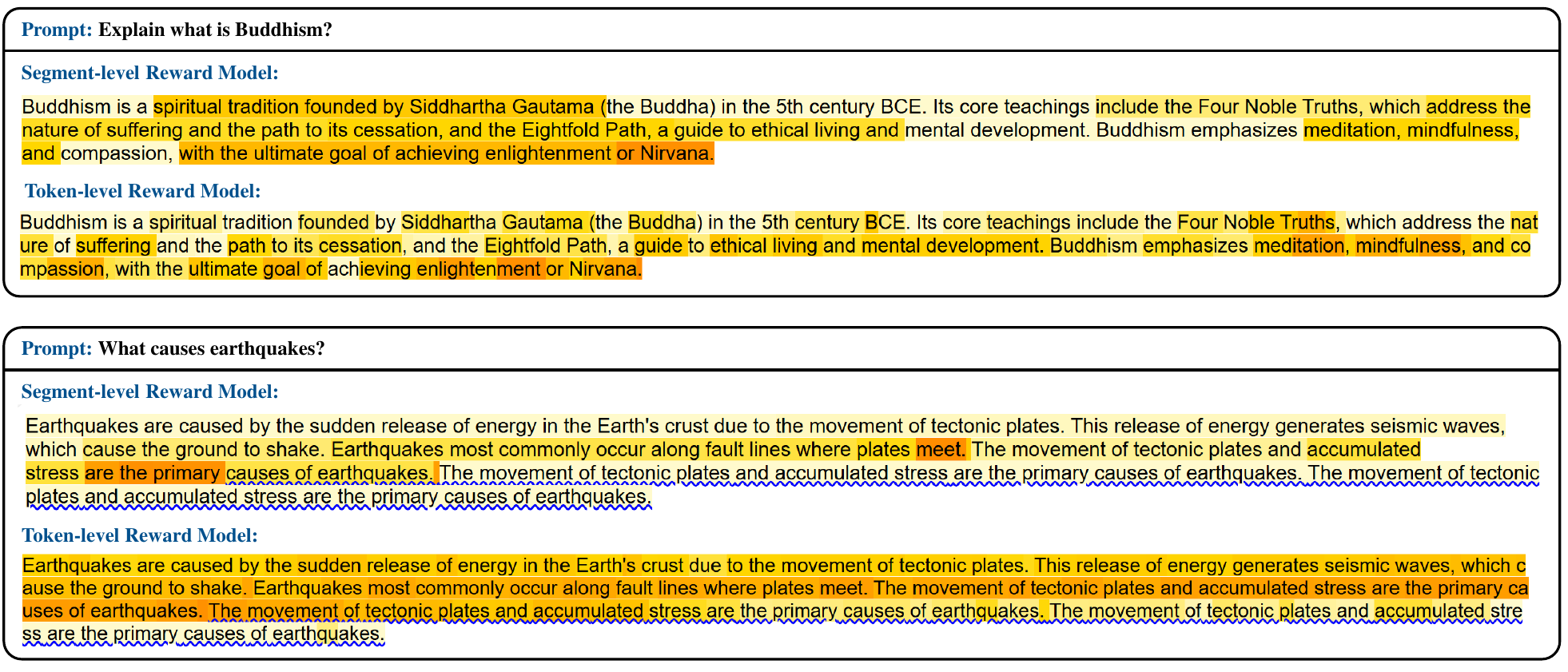} % width=8.5cm
\vspace{-2em}
\captionsetup{font=small}
\caption{
\small
Examples of dense reward assignment for text sequences encountered in PPO training, comparing our segment-level reward model and the recent token-level design on normal text (\textbf{{Top}}) and text with verbosity/repetition (\textbf{{Bottom}}).
Darker color indicates higher reward.
In the bottom half, repeated sentences are \uwave{underlined}.
}
 % \vspace{-1em}
\label{fig:seg_token_reward_comp}
\end{figure}

\cref{sec:example_responses} provides generation examples from our main LM policy.
\cref{table:openllm_main} in \cref{sec:additional_tab_results} compares the LM policies in \cref{table:diff_seg_method} on tasks on OpenLLM Leaderboard.
Both show that our method, while achieving strong RLHF training, does not suffer from the ``alignment tax'' \citep{askell2021general}.

\subsection{Ablation Study}\label{sec:exp_abla}
This section considers the following four research questions to better understand our method.
To save compute, all ablation studies are conducted on the 3.8B Phi3-mini Instruct used in \cref{table:diff_seg_method}.

\textbf{(a):} \textit{Can our segment-level reward model reasonably distinguish key segments?}

In \cref{fig:seg_token_reward_comp}, we qualitatively compare dense reward assignments from our segment-level reward model versus the recent token-level approach on two text sequences appeared in PPO training, representing normal text (\textit{Top}) and broken text with verbosity/repetition (\textit{Bottom}), where repetitions are underlined.

The color blocks in \cref{fig:seg_token_reward_comp} demonstrate that our entropy-based approach segments the text into meaningful semantic units, whereas a token may represent only part of a word.
Further, \cref{fig:seg_token_reward_comp} confirms that our segment-level reward model assigns higher rewards to  key segments in the responses.
Meanwhile, we see that the token-level model does not have consistent reward assignment over even a word --
often less understandably (only) highly rewards some letters of a word (\eg, ``Tr\uline{uth}s'', ``med\uline{itation}, comp\uline{assion}, ach\uline{ieving}'').
The benefit of our motivation of a semantically complete action space is further manifested by the example of repeated sentences in \cref{fig:seg_token_reward_comp} \textit{Bottom}, where our model assigns consistent low rewards to the repeated sentences, effectively refraining the LM from verbosity/repetition. 
By contrast, due to the semantic incompleteness of each action, the token-level reward model still assigns high rewards to tokens in the repetitions, even in the second repeat, which is undoubtedly undesirable.

\textbf{(b):} \textit{How will PPO training perform if we use different constructions of reward normalizers?}

\begin{table}[tb]
\captionsetup{font=small}
\caption{
\small
Comparison of different constructions of segment-level reward normalizers, on performance of the resulted PPO-trained LM policies.
Highest numeric of each metric is in bold.
} 
\label{table:normalization_methods}
\centering 
\vspace{-1em}
\resizebox{.95\textwidth}{!}
{
\small
\begin{tabular}{@{}lcccccccc@{}}
\toprule
\multirow{2}{*}{\begin{tabular}[c]{@{}l@{}}Reward \\ Normalizer\end{tabular}} & 
\multicolumn{3}{c}{AlpacaEval 2.0} & 
& \multicolumn{2}{c}{Arena-Hard} & 
& MT-Bench \\  
\cmidrule(l){2-4} \cmidrule(l){6-7} \cmidrule(l){9-9}
 & LC (\%) & WR (\%) & \# char & & WR (\%) & \# token & & GPT-4o \\ 
\midrule
No Reward Normalization & 19.64 & 24.10 & 2446 & & 29.9 & 625 & & 7.25 \\ 
Global Statistics of All & 17.34 & 22.11 & 2420 & & 31.3 & 639 & & 7.14 \\ 
Statistics of the Last Rewards & 20.30 & 24.72 & \textbf{2551} & & 29.2 & \textbf{641} & & 7.10 \\ 
\midrule
Regression-based (\cref{sec:method:ppo_policy}) & \textbf{31.05} & \textbf{34.53} & {2257} & & \textbf{34.0} & {593} & & \textbf{7.65} \\ 
\bottomrule
\end{tabular}
}
\vspace{-1em}
\end{table}

Recall that in our PPO training (\cref{sec:method:ppo_policy}), we use simple linear regression to fit location-aware mean and std functions that provide reward normalizers at arbitrary locations of the text sequence.
To study if this design is over-engineering, we compare our main method with three simpler constructions of segment-level reward normalizers: \textbf{\textit{(A)}} no reward normalization; \textbf{\textit{(B)}} using the scalar global mean and std over all segment-level rewards in the reward calibration dataset; and \textbf{\textit{(C)}} using the scalar mean and std over the last segment-level rewards in each response of the calibration set, mimicking the normalizers in the classical bandit approach.
\cref{table:normalization_methods} compares the resulted LM policies.

In \cref{table:normalization_methods}, we clearly see that normalizing (dense) reward  by improper reward statistics is akin to no reward normalization, as all three baselines have significantly lower benchmark scores that our regression-based approach and undesirable longer response lengths.
As discussed in details in \cref{sec:more_on_reward_norm}, the linguistic structure of the response leads to certain correlation between the mean and std of segment-level reward values and the normalized location of segment in the response, \eg, in the early or middle or later part.
This necessitates our design of location-aware reward normalizers that are able to capture the reward statistics at each arbitrary location of the sampled text sequence, since constant normalization statistics can be insufficient to properly normalize  the rewards of text segments at different parts of the text sequence, as verified in  \cref{table:normalization_methods}.
Future work may extend our linear regression-based normalizer functions in \cref{sec:method:ppo_policy} with non-linearity and/or more features.

\textbf{(c):} \textit{What will happen if we use different strategies for within-segment reward interpolation?}

\begin{table}[t]
\captionsetup{font=small}
\caption{
\small
Comparison of different within-segment reward interpolation strategies. Shown are the results of the resulted PPO-trained LM policies on AlpacaEval 2.0 and Arena-Hard.
Highest numeric of each metric is in bold.
} 
\label{table:diff_interp_method}
\centering 
\vspace{-1em}
\resizebox{.95\textwidth}{!}
{
\small
\begin{tabular}{@{}lcccccccc@{}}
\toprule
\multirow{2}{*}{\begin{tabular}[c]{@{}l@{}}Interpolation \\ Strategy\end{tabular}} & 
\multicolumn{3}{c}{AlpacaEval 2.0} & 
& \multicolumn{2}{c}{Arena-Hard} & 
& MT-Bench \\  
\cmidrule(l){2-4} \cmidrule(l){6-7} \cmidrule(l){9-9}
 & LC (\%) & WR (\%) & \# char & & WR (\%) & \# token & & GPT-4o \\ 
\midrule
No Interpolation & 25.98 & 34.53 & \textbf{2666} & & \textbf{39.1} & \textbf{675} & & 7.45 \\ 
Repeat Segment Reward & 26.34 & 23.48 & 1795 & & 23.0 & 512 & & 7.42 \\ \midrule
Even Split (\cref{sec:method:ppo_policy}) & \textbf{31.05} & \textbf{34.53} & 2257 & & 34.0 & 593 & & \textbf{7.65} \\ 
\bottomrule
\end{tabular}
}
\vspace{-1em}
\end{table}

Recall from \cref{sec:method:ppo_policy} that, to further densify the learning signal in RLHF for enhancing training, we interpolate the segment-level rewards by evenly splitting the reward of a segment  to each of its constituting token.
We now compare this even-split interpolation strategy with two other intuitive alternatives: \textbf{\textit{(A)}} no interpolation on the segment-level rewards, use $0$ for technical padding in PPO (``No Interpolation''); \textbf{\textit{(B)}} repeat the segment-level reward of a segment to each token in it (``Repeat Segment Reward'').
\cref{table:diff_interp_method} shows the performance of the resulted PPO-trained LM policies.

In conjunction with our main result \cref{table:diff_seg_method}, in \cref{table:diff_interp_method}, we see that these two alternatives still provide (relatively) effective RLHF training on Phi3.1-mini Instruct, in reference to the results of the classical bandit approach in \cref{table:diff_seg_method}.
Nevertheless, we see that the generation length from ``No Interpolation'' is significantly longer, while ``Repeat Segment Reward'' is too short.
Probing into those long text sequences encountered in PPO training, we found that they typically contain some very negative segment-level rewards that refrains the behavior of long generation from being learned by the policy LM.
Such very negative reward signals may be diluted by the technical zero-padding in ``No Interpolation'', leading to overly long text generation, whereas they are overly amplified in ``Repeat Segment Reward'', resulting in too-strong punishment for long texts and hence too-short generations. 
By contrast, the even-split interpolation strategy in our main method provides densified learning signal of a proper scale, which we attribute to the implicit (segment-) length normalization inherited from the operation of dividing by segment length in an even split. % 
Future work may design a proper non-even split of segment-level reward over each token in the text segment.

\textbf{(d):} \textit{With a different entropy cutoff $\cent$ for text segmentation, how will our method perform?}

% ----------------------------------------------------
\begin{figure}[t]
     % \vspace{-0.6em}
     \centering
\includegraphics[width=.52\textwidth]{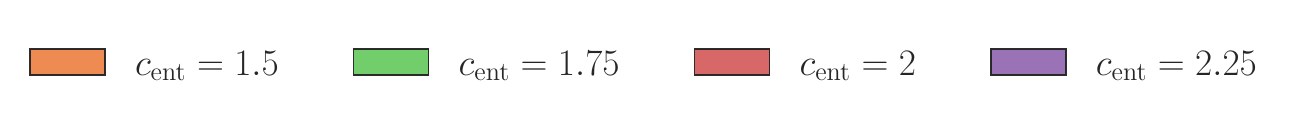}
\\
% \vspace{-1.0mm}
     % \hspace{1em}
     \begin{subfigure}[b]{0.19\textwidth}
         \centering
         \includegraphics[width=\textwidth]{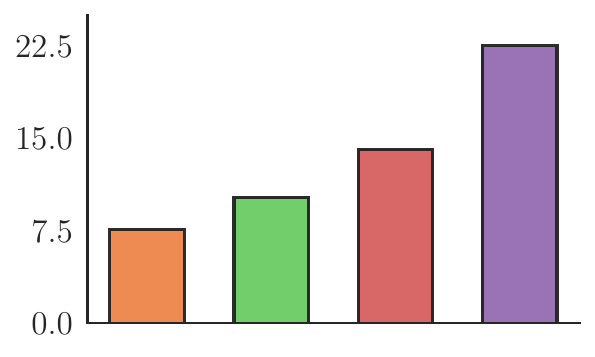}
         \captionsetup{font=scriptsize}
         \vspace{-6mm}
         \caption{\scriptsize{Avg. Seg. Len (token)}}
         \label{fig:seg_len}
     \end{subfigure}
     \hfill
     \begin{subfigure}[b]{0.19\textwidth}
         \centering
         \includegraphics[width=\textwidth]{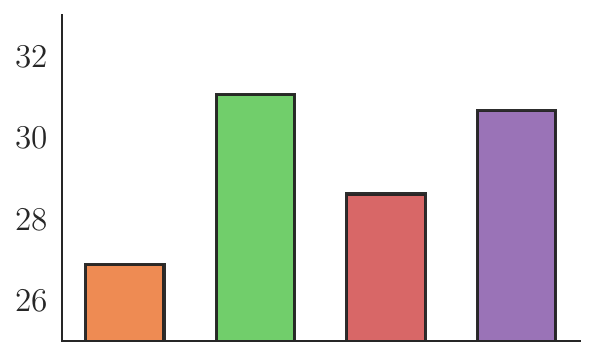}
         \captionsetup{font=scriptsize}
         \vspace{-6mm}
         \caption{\scriptsize{AlpacaEval 2 (LC)}}
         \label{fig:ent_alpaca2_lc}
     \end{subfigure}
     \hfill
     % \hspace{1em}
    %  \rulesep
    \begin{subfigure}[b]{0.19\textwidth}
         \centering
         \includegraphics[width=\textwidth]{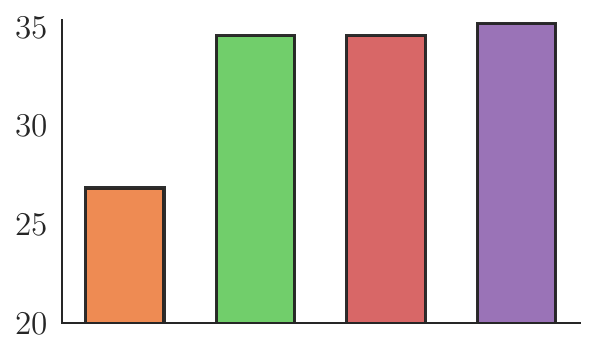}
         \captionsetup{font=scriptsize}
         \vspace{-6mm}
         \caption{\scriptsize{AlpacaEval 2 (WR)}}
         \label{fig:ent_alpaca2_wr}
     \end{subfigure}
     \hfill
     % \hspace{1em}
    %  \rulesep
    \begin{subfigure}[b]{0.19\textwidth}
         \centering
         \includegraphics[width=\textwidth]{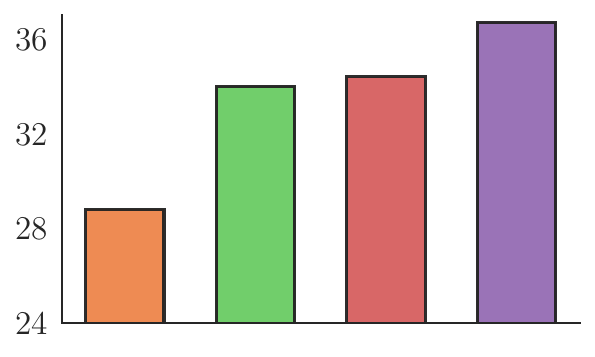}
         \captionsetup{font=scriptsize}
         \vspace{-6mm}
         \caption{\scriptsize{Arena-Hard}}
         \label{fig:ent_arenahard}
     \end{subfigure}
     \hfill
     % \hspace{1em}
    %  \rulesep
    \begin{subfigure}[b]{0.19\textwidth}
         \centering
         \includegraphics[width=\textwidth]{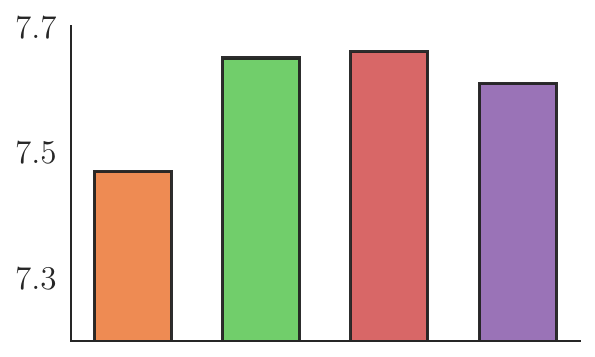}
         \captionsetup{font=scriptsize}
         \vspace{-6mm}
         \caption{\scriptsize{MT-Bench}}
         \label{fig:ent_mtbench}
     \end{subfigure}
     \vspace{-2mm}
     \captionsetup{font=small}
        \caption{ 
        \small
        Performance comparison among different entropy cutoffs $\cent$ for entropy-based text segmentation, comparing PPO-trained LM policy's benchmark scores and average segment length (``Avg. Seg. Len'') in terms of number of tokens.
        % The reward model and corresponding policy are both trained under the same $\cent$ as specified.
        % For reward models' performance, we plot the average accuracy over the entire RewardBench.
        }
        \label{fig:ent_cutoff}
        \vspace{-1em}
\end{figure}
% ----------------------------------------------------

As discussed in \cref{sec:exp_setup}, for main results, we use entropy cutoff $\cent = 1.75$ for entropy-based text segmentation.
To investigate the impact of $\cent$, 
in \cref{fig:ent_cutoff}, we vary the value of $\cent \in \{1.5, 1.75, 2.0, 2.25\}$, and compare the performance of the resulted PPO-trained LM policies as well as the average segment length of the PPO-trained LM policy.

As seen in \cref{fig:ent_cutoff}, similar to the discussion of token-level approach  in \cref{sec:intro}, a smaller $\cent = 1.5$, which chops text sequence into finer pieces with smaller average segment length, may result in semantically less complete segments, leading to less accurate reward modeling and the subsequent weaker LM policy.
A reasonably larger entropy cutoff, such as $\cent \in [1.75, 2.25]$ that corresponds to an average segment length of $10$ to $22$ in \cref{fig:seg_len} (or about $3$ to $7$ words), leads to much better PPO-trained LMs.
This coincides with the advantage of our segment-level design over the prior token-level design in Tables~\ref{table:diff_seg_method}-\ref{table:action_definitions_llama3} and verifies our goal of a more semantically complete action space.

\section{Conclusion and Limitations}
In this paper, we propose to train and utilize a segment-level reward model for improved RLHF in LMs, motivated by both a denser reward signal in RL-based LM training and semantic completeness of each action for accurate reward assignment. 
Our method and insight are validated through extensive experiments, ablation studies, and backbone models of different sizes. 
Nevertheless, as an early effort in reconsidering the action space in RLHF, our experiments are currently confined to PPO training on a free-form dialog-alike dataset and instruction-following benchmark evaluations. 
Our future work includes scaling up to even larger LMs, testing  our method on other types of tasks such as math problem solving and code generation, and applying to other RL algorithms, such as REINFORCE.

\bibliographystyle{plainnat}
\bibliography{arxiv}

\newpage
\appendix
\section{Algorithm Box}\label{sec:algo_box}
\cref{algo:main_method} summarizes our method in \cref{sec:method} on training the segment-level reward model and utilizing it in PPO-based RLHF LM training.
Note that all operations in \cref{algo:main_method} can be efficiently conducted in batch mode, parallel for multiple sample points at once. 

% \vspace{-3em}
\setlength{\textfloatsep}{0.1cm}
\setlength{\floatsep}{0.1cm}
\begin{algorithm}[H]
% \captionsetup{font=small}
\caption{Training and Utilizing Our Segment-level Reward.}
\label{algo:main_method}
\begin{algorithmic}
\STATE \textbf{Input:} 
Binary preference dataset $\gD_{\mathrm{pref}} = \{(x, y^w, y^l)\}$ for reward model training,
prompt set $\gD_{\mathrm{pol}} = \{x\}$ for policy learning,
supervised fine-tuned model $\pisft$,
reward model training steps $M_\mathrm{rew}$,
LM policy training steps $M_\mathrm{pol}$,
entropy cutoff $\cent$, KL coefficient $\beta$ for RLHF PPO training.
\STATE

% \STATE
\STATE \textbf{Initialization:} Initialize the segment-level reward model $r_\phi$ and LM policy $\pi_\theta$ from $\pisft$,
fix the aggregation function $f(\cdot)$ as the Average in \cref{eq:param_seq_eval},
initialize other components in the off-the-shelf RLHF PPO routine as specified.
\STATE

\STATE \textcolor{gray}{// Training the segment-level reward model}
\STATE Use $\pisft$ and $\cent$ to split the responses $\{(y^w, y^l)\}$ in $\gD_{\mathrm{pref}} = \{(x, y^w, y^l)\}$ into segments.
\FOR{$\mathrm{iter} \in \{1, \ldots,M_\mathrm{rew}\}$}
\STATE Sample a minibatch $\gB =  \{(x_i, y^w_i, y^l_i)\}_{i} \sim \gD_{\mathrm{pref}}$.
\STATE With $f(\cdot)$ and $\tau$, calculate  $e_\phi(x_i, y^{w}_i)$ and  $e_\phi(x_i, y^{l}_i)$ by \cref{eq:param_seq_eval} for $(x_i, y^w_i, y^l_i) \in \gB$.
\STATE Optimize reward model $r_\phi$ by \cref{eq:seg_bt}.
\ENDFOR
\STATE

\STATE \textcolor{gray}{// Utilizing the segment-level reward model in PPO-based LM policy learning}
\STATE Estimate the reward normalizer functions $\mathrm{Mean}(p)$ and $\mathrm{Std}(p)$ as described in \cref{sec:method:ppo_policy}.
\FOR{$\mathrm{iter} \in \{1, \ldots,M_\mathrm{pol}\}$}
\STATE Sample a minibatch $\gB =  \{x_i\}_{i} \sim \gD_{\mathrm{pol}}$.
\STATE Sample a response $y_i \sim \pi_\theta(\cdot \given x_i)$ for each $x_i \in \gB$
\STATE Use $\pisft$ and $\cent$ to segment each $y_i$; record the completion portion $p$ of each segment.
\STATE Use $r_\phi$ to assign a segment-level reward to each segment $a_t$ in each $y_i$
\STATE Normalize each segment reward $r_\phi(s_t, a_t)$ as $r_\phi(s_t, a_t) \leftarrow (r_\phi(s_t, a_t) - \mathrm{Mean}(p))/\mathrm{Std}(p)$.
\STATE Interpolate $r_\phi(s_t, a_t)$ to each token $y_i$, as $\forall\, a_t \in y, \forall\, y_i \in a_t, \tilde r_\phi([x, y_{<i}], y_i) = r_\phi(s_t, a_t) / |a_t|$
\STATE With KL coefficient $\beta$, optimize policy LM $\pi_\theta$ against $\tilde r_\phi$ by the PPO routine.
\ENDFOR
\end{algorithmic}
\end{algorithm}
\setlength{\textfloatsep}{0.2cm}
\setlength{\floatsep}{0.1cm}
% \vspace{-1em}

\section{Additional Results} \label{sec:additional_results}

\subsection{Tabular Results} \label{sec:additional_tab_results}

\begin{table}[H]
\captionsetup{font=small}
\caption{
\small
Evaluation results of downstream tasks on the HuggingFace OpenLLM Leaderboard  \citep{beeching2023open}, comparing LM policies in
\cref{table:diff_seg_method}.
} 
\label{table:openllm_main}
\centering 
\vspace{-1em}
\resizebox{1\textwidth}{!}
{
\begin{tabular}{@{}lcccccc|c@{}}
\toprule
Action Definition       & ARC   & TruthfulQA & Winograd & HellaSwag & MMLU  & GSM8K & Average \\ \midrule
Phi-Instruct                            & \textbf{64.76} & 54.44      & 74.51    & 79.03     & 70.41 & 81.6  & 70.79   \\ \midrule
Bandit (Sequence)                        & \textbf{64.76} & \textbf{55.11}      & 74.35    & 79.32     & 70.42 & 77.8  & 70.29   \\
Sentence                                 & 63.40 & 53.99      & 72.93    & 79.34     & 70.42 & 84.1  & 70.70   \\
Token                                    & 62.71 & 53.94      & 71.43    & \textbf{79.46}     & \textbf{70.55} & \textbf{87.3}  & 70.90 \\ \midrule
Segment (\textbf{Ours})    & 62.71 & 54.74      & 72.06    & 79.23     & 70.42 & 86.7  & \textbf{70.98}  \\ \midrule
Bandit as Segment                        & 64.16 & 54.62      & 74.66    & 78.95     & \textbf{70.55} & 81.0  & 70.66   \\
Segment as Bandit                        & 64.33 & 54.81      & \textbf{74.74}    & 79.23     & 70.39 & 78.6  & 70.35  \\ \bottomrule
\end{tabular}
}
\end{table}

% \newpage
\subsection{Examples of Generated Responses} \label{sec:example_responses}

Below, we provide examples of generated responses from our LM policy in the main results (\cref{table:diff_seg_method}) on the benchmarks AlpacaEval 2.0 (first two examples) and Arena-Hard (last two examples).

\vspace{1em}

\begin{mdframed}
\textbf{Prompt:} When are hops added to the brewing process?

\textbf{Response:} Hops, the flowers of the hop plant \textit{Humulus lupulus}, are a crucial ingredient in brewing beer, primarily used for their bittering, flavoring, and aromatic properties. They are typically added to the brewing process at various stages, depending on the desired outcome in terms of bitterness, flavor, and aroma. The timing of hop addition is a critical aspect of brewing that significantly influences the final beer's character.

1. \textbf{Boiling Stage (Early Boil)}: This is the most common stage for adding hops. Adding hops early in the boil (usually within the first 15 minutes) primarily contributes to the beer's bitterness. The longer the hops are boiled, the more bitterness is extracted due to the breakdown of alpha acids, which are responsible for this bitterness. This process is known as isomerization. Early boil hops can also contribute some flavor, but the impact on aroma is minimal because the volatile compounds responsible for aroma are largely evaporated during this long boil.

2. \textbf{Middle of the Boil}: Adding hops in the middle of the boil (around 15 to 30 minutes) can strike a balance between flavor and aroma. The bitterness level is moderate, and the hop flavors and some aromatic compounds are preserved better than in the early boil, thanks to the shorter exposure time.

3. \textbf{Late Boil (Last 10-15 minutes)}: Hops added late in the boil contribute significantly to the beer's aroma and flavor, with minimal to no contribution to bitterness. This is because the shorter boiling time allows the volatile aromatic compounds to remain intact, while the alpha acids responsible for bitterness are less extracted. This technique is often used in styles that emphasize hop aroma and flavor, such as India Pale Ales (IPAs) and New England IPAs (NEIPAs).

4. \textbf{Fermentation Stage}: Some brewers also add hops during the fermentation process, a technique known as dry hopping. This is done after the primary fermentation has completed, and the beer is transferred to a secondary fermenter or directly to the bottle/keg. Dry hopping enhances the beer's aroma without significantly affecting its bitterness. The extended contact time allows the hop oils to infuse the beer, imparting a fresh, fruity, and floral aroma.

5. \textbf{Post-Fermentation (Aging)}: In some cases, especially with barrel-aged beers, hops may be added during the aging process to impart additional flavor and aroma characteristics. This can be a more experimental approach, as the interaction between the hops and the aging environment (e.g., oak barrels, different beer styles) can lead to unique and complex profiles.

The choice of when to add hops is a critical decision that brewers make based on the style of beer they are aiming to produce and the specific flavor, aroma, and bitterness profiles they desire. Experimentation and experience play significant roles in fine-tuning this aspect of the brewing process to achieve the perfect balance for each batch.
\end{mdframed}

% \newpage
\begin{mdframed}
\textbf{Prompt:} Introduce the “financial markets and institutions“ by Frederic S. Mishkin.

\textbf{Response:} In his comprehensive work, \textit{The Economics of Money, Banking, and Financial Markets}, Frederic S. Mishkin delves into the intricate world of financial markets and institutions, providing a thorough examination of their roles, functions, and significance in the global economy. The book, published in 2000, is a well-regarded resource for students, professionals, and policymakers alike, as it explores the complexities of financial systems and their impact on economic growth, stability, and development.

Financial markets and institutions are integral components of the broader economic system, serving as the backbone for the efficient allocation of resources, risk management, and the facilitation of transactions between various participants. Mishkin's book offers a deep dive into the various types of financial markets, such as capital markets, money markets, and foreign exchange markets, as well as the institutions that operate within these environments, including banks, investment firms, and regulatory bodies.

The book begins by establishing a solid foundation in the basic concepts and principles of money, banking, and financial markets, ensuring that readers have a clear understanding of the fundamental ideas that underpin the entire field. Mishkin then progresses to explore the role of financial intermediaries, such as banks and other institutions, in connecting savers and borrowers, and how these entities contribute to the overall efficiency and stability of the financial system.

One of the key themes in Mishkin's work is the importance of financial markets and institutions in promoting economic growth and development. He examines the ways in which these entities facilitate the flow of funds from savers to productive investments, thereby supporting innovation, entrepreneurship, and job creation. Furthermore, Mishkin discusses the role of financial markets in providing risk management tools, such as derivatives and insurance products, which help individuals and businesses protect themselves against potential losses.

Another crucial aspect of Mishkin's exploration is the examination of the regulatory framework that governs financial markets and institutions. The author emphasizes the need for effective oversight and regulation to ensure the stability of the financial system, protect consumers, and prevent financial crises. Throughout the book, Mishkin provides real-world examples and case studies to illustrate the consequences of both well-functioning and poorly regulated financial markets, highlighting the importance of maintaining a balanced and robust regulatory environment.

In addition to discussing the theoretical aspects of financial markets and institutions, Mishkin also addresses practical considerations, such as the impact of technology on the financial industry, the role of central banks in monetary policy, and the challenges posed by globalization and financial integration. By offering a comprehensive and well-rounded perspective on the subject, Mishkin's work serves as an invaluable resource for anyone seeking to understand the complexities and nuances of financial markets and institutions in the modern economic landscape.
\end{mdframed}

\newpage

\begin{mdframed}
\textbf{Prompt:} Provide me with a breakfast recipe that is quick to make and is high in protein (at least 30 grams) and has a variety of ingredients.

\textbf{Response:} Certainly! Here's a quick and delicious high-protein breakfast recipe called "Spinach and Feta Egg Muffins." This recipe is not only packed with protein but also incorporates a variety of ingredients, including spinach, feta cheese, and eggs. It's perfect for a healthy, on-the-go breakfast or a satisfying meal prep option.

\textbf{Ingredients:}
\begin{itemize}
    \item 6 large eggs
    \item 1 cup fresh spinach, chopped
    \item 1/2 cup feta cheese, crumbled
    \item 1/4 cup red onion, finely chopped
    \item 1/4 cup cherry tomatoes, halved
    \item 1/4 cup green bell pepper, finely chopped
    \item 1/4 cup mushrooms, finely chopped
    \item 1/4 cup fresh parsley, chopped
    \item Salt and pepper, to taste
    \item 1/4 teaspoon garlic powder (optional)
    \item 1/4 teaspoon paprika (optional)
    \item 1/4 teaspoon dried oregano (optional)
\end{itemize}

\textbf{Instructions:}
\begin{enumerate}
    \item Preheat your oven to 375°F (190°C) and lightly grease a muffin tin with cooking spray or a small amount of oil.
    \item In a large mixing bowl, whisk the eggs until well combined.
    \item Add the chopped spinach, feta cheese, red onion, cherry tomatoes, green bell pepper, mushrooms, and parsley to the egg mixture. Stir until all the ingredients are evenly distributed.
    \item Season the mixture with salt, pepper, garlic powder, paprika, and dried oregano (if using) to taste.
    \item Pour the egg mixture into the prepared muffin tin, filling each cup about 3/4 full.
    \item Place the muffin tin in the preheated oven and bake for 20-25 minutes, or until the egg muffins are set and the tops are lightly golden.
    \item Remove the muffin tin from the oven and let the egg muffins cool for a few minutes before transferring them to a wire rack to cool completely.
    \item Once cooled, you can store the egg muffins in an airtight container in the refrigerator for up to 3-4 days or freeze them for up to 3 months.
    \item To reheat, simply place the egg muffins in a microwave-safe dish and heat for 15-20 seconds, or until warmed through.
\end{enumerate}

Enjoy your high-protein, delicious, and nutritious Spinach and Feta Egg Muffins!
\end{mdframed}

% \newpage

\begin{mdframed}
\textbf{Prompt:} Write a python function that solves a quadratic equation.

\textbf{Response:} To solve a quadratic equation of the form \(ax^2 + bx + c = 0\), we can use the quadratic formula:

\[
x = \frac{-b \pm \sqrt{b^2 - 4ac}}{2a}
\]

This formula gives us the roots of the quadratic equation, where \(a\), \(b\), and \(c\) are coefficients of the equation, and \(x\) represents the variable. The term under the square root, \(b^2 - 4ac\), is known as the discriminant. The discriminant determines the nature of the roots:

- If the discriminant is positive, there are two distinct real roots.
- If the discriminant is zero, there is exactly one real root (also known as a repeated or double root).
- If the discriminant is negative, there are no real roots, but two complex roots.

Here's a Python function that calculates the roots of a quadratic equation, handling all three cases:

\begin{lstlisting}[language=Python]
import cmath

def solve_quadratic(a, b, c):
    """
    Solves a quadratic equation of the form ax^2 + bx + c = 0.

    Parameters:
    - a: Coefficient of x^2
    - b: Coefficient of x
    - c: Constant term

    Returns:
    - A tuple containing the roots of the equation. If the roots are complex,
      they are returned as complex numbers.
    """
    # Calculate the discriminant
    discriminant = cmath.sqrt(b**2 - 4*a*c)
    
    # Calculate the two roots using the quadratic formula
    root1 = (-b + discriminant) / (2*a)
    root2 = (-b - discriminant) / (2*a)
    
    return (root1, root2)

# Example usage
a, b, c = 1, 5, 6  # Coefficients for the equation x^2 + 5x + 6 = 0
roots = solve_quadratic(a, b, c)
print(f"The roots of the equation are: {roots[0]} and {roots[1]}")

# For a complex discriminant
a, b, c = 1, 2, 5  # Coefficients for the equation x^2 + 2x + 5 = 0
roots = solve_quadratic(a, b, c)
print(f"The roots of the equation are: {roots[0]} and {roots[1]}")
\end{lstlisting}

This function uses the `cmath` module to handle complex square roots, ensuring that it can return complex roots when necessary. The function calculates and returns both roots, which can be real or complex numbers, depending on the discriminant.
\end{mdframed}

% \newpage

\section{More Implementation Details} \label{sec:exp_details}

\begin{tabular}{cc}
    \begin{minipage}{.48\linewidth}
        \begin{table}[H]
        \captionsetup{font=small}
        \caption{
        \small Model hyperparameters used in reward model training.
        } 
        \label{table:rm_train_hyperparameters} 
        \centering 
        \begin{tabular}{@{}ll@{}}
        \toprule
        Hyperparameter              & Value   \\ \midrule
        Batch Size          & 128     \\
        Training Epochs                 & 1       \\
        Max prompt length            & 1728    \\
        Max total length                    & 2048    \\
        DeepSpeed ZeRO stage             & 3       \\
        Optimizer                   & Adam    \\
        Learning Rate              & 1e-6    \\
        Gradient clipping norm          & 1.0     \\
        Entropy threshold $\cent$         & 1.75     \\
        \\
        \\
    \\\\
        \bottomrule
        \end{tabular}
        \end{table}
    \end{minipage} & 
    \begin{minipage}{.48\linewidth}
        \begin{table}[H]
        \captionsetup{font=small}
        \caption{
        \small Model hyperparameters used in PPO-based LM policy training.
        } 
        \label{table:ppo_train_hyperparameters} 
        \centering 
        % \resizebox{\textwidth}{!}{
        \begin{tabular}{@{}lll@{}}
        \toprule
        Hyperparameter            & Phi-mini & Llama-3-8B \\ \midrule
        Batch Size                & 128      & 128        \\
        Rollout batch size        & 1024     & 1024       \\
        Micro rollout batch size  & 16       & 4          \\
        Training Epochs           & 2        & 1          \\
        Max prompt length         & 1024     & 1024       \\
        Max generation length     & 1024     & 1024       \\
        DeepSpeed ZeRO stage      & 2        & 2          \\
        Actor learning rate       & 5e-7     & 5e-7       \\
        Critic learning rate      & 9e-6     & 9e-6       \\
        Gradient clipping norm    & 1.0      & 1.0        \\
        Entropy threshold $\cent$ & 1.75     & 2.0        \\
        Value clipping            & 0.25     & 0.25       \\
        KL coefficient $\beta$    & 0.01     & 0.01       \\ \bottomrule
        \end{tabular}
        % }
        \end{table}
    \end{minipage}
\end{tabular}

\vspace{14pt}

\myparagraph{Implementation Details.} 
 We tabulate detailed parameter settings in \cref{table:rm_train_hyperparameters} and \cref{table:ppo_train_hyperparameters}. 
 Most of them are the same as the default setting in \href{https://github.com/OpenRLHF/OpenRLHF}{OpenRLHF}.
  Both the reward model and PPO training employ the Adam optimizer \citep{adam2014}, with \(\beta_1 = 0.9\) and \(\beta_2 = 0.95\). To save GPU memory, we use gradient checkpointing \citep{chen2016training} and flash attention \citep{dao2205fast}.

For reward model training, we set the maximum prompt sequence length as 1792 tokens, with the total sequence length (including both prompt and response) capped at 2048 tokens. 
During data preprocessing, we apply left truncation to the prompt and right truncation to the response. 
If the EOS token in the response is truncated, we manually change the last token in the truncated response to the EOS token. 
The global mini batch size for reward model training is set to 128, with each GPU processing a micro batch size of 8. 
To facilitate distributed training, we utilize \href{https://github.com/microsoft/DeepSpeed}{DeepSpeed ZeRO-3}. 
For our segment-level reward model, the entropy threshold is set to \(\cent = 1.75\) for training with the Phi-series models and \(\cent = 2\) for the Llama-3-8B model.
The baseline bandit reward model is technically implemented as  setting the entropy threshold $\cent = 1000$, restricting reward computation to the EOS token only, while the baseline token-level reward model is implemented as setting the entropy threshold $\cent = 0$, ensuring that a reward is computed for each token in the text sequence.

For PPO training, the replay buffer size (\texttt{rollout\_batch\_size}) is set to 1024, while the batch size per GPU for generation (\texttt{micro\_rollout\_batch\_size}) is configured as 16 for Phi-mini and 4 for Llama-3-8B. The maximum prompt sequence length is set as 1024 tokens, and the maximum generated sequence length is also set to 1024 tokens. 
In PPO's on-policy sampling, for each prompt in the mini-batch, a single response is sampled via top-$p$ sampling with $p=1.0$ and sampling temperature 1.0. We use DeepSpeed ZeRO-2 for distributed training. 
The actor learning rate is set to the default value of \(5 \times 10^{-7}\), and the critic learning rate is also the default value of \(9 \times 10^{-6}\). The  clipping coefficient for value loss (value clip) is set to 0.25 for PPO training based on segment- and token-level reward model, and as default to 0.2 for bandit-reward-based PPO training. The clipping coefficient for policy loss  (eps clip) is set to 0.2. The KL coefficient is kept to the default value of $\beta=0.01$.

\newpage
\myparagraph{Chat Template.}
Below is the chat template used for PPO training, reward model training, and evaluation sampling with Phi-3. The original Phi-3-Mini chat template includes an additional EOS token, which we removed in experiments due to incompatibility, leaving only the \texttt{<|end|>} token.

\begin{mdframed}

\vspace{8pt}

\textless$|$user$|$\textgreater

\noindent \{the user query\} \textless$|$end$|$\textgreater

\noindent \textless$|$assistant$|$\textgreater

\noindent \{Assistant Response\} \textless$|$end$|$\textgreater

\end{mdframed}

For Llama-3, the chat template is as follows:

\begin{mdframed}

\vspace{8pt}

\textless$|$begin\_of\_text$|$\textgreater \textless$|$start\_header\_id$|$\textgreater user\textless$|$end\_header\_id$|$\textgreater
\vspace{8pt}

\noindent \{the user query\}\textless$|$eot\_id$|$\textgreater \textless$|$start\_header\_id$|$\textgreater assistant\textless$|$end\_header\_id$|$\textgreater
\vspace{8pt}

\noindent \{Assistant Response\}\textless$|$eot\_id$|$\textgreater

\end{mdframed}

\section{Computation of Location-Aware Reward Normalizers via Regression}\label{sec:more_of_linear_fit_function}

The computation process for the location-aware reward normalizers involves several key steps, starting with data sampling and culminating in regression modeling for the estimation of reward normalization functions.

First, 60,000 data points are randomly sampled from the \texttt{Preference-700K} dataset, which includes pairs of prompts, chosen responses, and rejected responses. Each response is processed by a segment reward model, where the segments within the response are indexed by their respective normalized location percentages. Specifically, the normalized location \( p \in (0, 1] \) is computed for each segment \( a_t \) as \( p = \frac{t}{T} \), where \( t \) is the index of the segment within the sequence and \( T \) represents the total number of segments in the response. The model then records the reward for each segment, producing a set of data points that consist of the segment's normalized location and its corresponding reward.

Next, this collected data is stored in a dictionary format, which is then saved as a JSON file. The data is subsequently loaded for further analysis, where the normalized locations (denoted as \( p \)) are sorted and the corresponding mean and standard deviation of rewards at each location are extracted.

To estimate the relationship between the normalized location and the reward statistics, we employ a linear regression approach using the \texttt{HuberRegressor} from the \texttt{sklearn} library, which is robust to outliers. We perform the regression on the log-transformed normalized locations, \( \log(p) \), to model the dependence of the mean reward \( \mu_p \) and the standard deviation \( \sigma_p \) of rewards at each location. The regression formulas are given by:

\[
\mathrm{Mean}(p) = w_\mu \times \log(p) + b_\mu
\]
\[
\mathrm{Std}(p) = w_\sigma \times \log(p) + b_\sigma
\]

Here, \( w_\mu \) and \( b_\mu \) are the regression coefficients for the mean reward, and \( w_\sigma \) and \( b_\sigma \) are those for the standard deviation. These coefficients are computed via closed-form solutions derived from the linear regression process.

Once the regression coefficients are obtained, we use them to compute the normalized rewards for each segment during the reinforcement learning process. The normalized reward \( r_\phi(s_t, a_t) \) is adjusted according to the location-aware normalizers:

\[
r_\phi(s_t, a_t) \leftarrow \frac{r_\phi(s_t, a_t) - \mathrm{Mean}(p)}{\mathrm{Std}(p)}
\]

\section{More on the Reward Normalizers in PPO Training}\label{sec:more_on_reward_norm}

{
To center the assigned rewards from the reward model and reduce their variance,
in most open-source (bandit) RLHF PPO implementations \citep[\eg,][]{havrilla-etal-2023-trlx,hu2024openrlhf}, the bandit reward of the newly sampled response $y$ is first ``Z-score'' normalized, before being fed into the PPO routine.
Concretely, for the prompt $x$ and sampled response $y$, the bandit reward $r_\phi(x, y)$ is normalized as
$r_\phi(x, y) \leftarrow (r_\phi(x, y) - \mu) / \sigma$, where $\mu$ and $\sigma$ are respectively the mean and standard deviation of (bandit) rewards in the reward calibration dataset.
The PPO routine starts by using this normalized $r_\phi(x, y)$, \eg, first subtract it by the KL regularization, and then calculate the advantage estimates and value function training target, \textit{etc}.
}

{
For the segment-level action space, we will then need to normalize the reward $r_\phi(s_t, a_t)$ for each segment $a_t$.
As shown in \cref{table:normalization_methods} (``Global Statistics of All''), the most intuitive idea of simply using the global mean and standard deviation over all segment-level rewards in the reward calibration dataset does not train a good LM.
Looking into the responses sampled in PPO training and in the reward calibration dataset, we find that, for example, the beginning segments of the responses are typically greeting alike phrases that are less informational and/or essential to respond to the given prompt, which tend to receive relatively lower rewards.
If we normalize the segment-level rewards of those early segments by the global mean and standard deviation, those normalized rewards will be significantly negative, rather than centered around $0$.
This will undesirably refrain the generation of necessary greeting alike phrases, resulting in an ``impolite LM'' and thus inferior benchmark results.
More generally, the linguistic structure of the response leads to certain correlation between the mean and standard deviation of segment-level reward values and the normalized location of segment in the response, \eg, in the early or middle or later part.
This observation motivates us to design location-aware reward normalizers that can approximately capture the reward statistics at an arbitrary location of the response, so that the normalized segment-level rewards can be more centered and less varying.
It is important to have proper reward normalizers at an \textit{arbitrary} location of the response, because the  response sampled in PPO training will have a stochastic total length, nondeterministic number of segments, and  less-controllable length of each segment.
These considerations motivate our design of the regression-based reward normalizer functions in \cref{sec:method:ppo_policy}.
}

\end{document}